\pdfoutput=1
\documentclass[11pt]{article}

\usepackage[]{acl}

\usepackage{times}
\usepackage{latexsym}

\usepackage[T1]{fontenc}
\usepackage{amsfonts,amsmath,amssymb,amsthm}
\usepackage{booktabs}
\usepackage{multirow}

\usepackage[utf8]{inputenc}

\usepackage{microtype}

\usepackage{inconsolata}

\usepackage{graphicx}

\title{Learning from Diverse Reasoning Paths with Routing and Collaboration}

\author{
    Zhenyu Lei$^{\blacklozenge}$ \:
    Zhen Tan$^\Diamond$\:  Song Wang$^\blacklozenge$ \\ \bf
    Yaochen Zhu$^\blacklozenge$ \:
    Zihan Chen$^\blacklozenge$ \:
    Yushun Dong$^\heartsuit$ \:
    Jundong Li$^\blacklozenge$ \\
    $^\blacklozenge$University of Virginia, $^\Diamond$Arizona State University, $^\heartsuit$Florida State University\\
    \texttt{\{vjd5zr, sw3wv, uqp4qh, brf3rx, jundong\}@virginia.edu}\\ \texttt{ztan36@asu.edu, \texttt{yd24f@fsu.edu}
} \\
}

\begin{document}
\maketitle
\begin{abstract}
Advances in large language models (LLMs) significantly enhance reasoning capabilities but their deployment is restricted in resource-constrained scenarios. Knowledge distillation addresses this by transferring knowledge from powerful teacher models to compact and transparent students.
However, effectively capturing the teacher's comprehensive reasoning is challenging due to conventional token-level supervision's limited scope. Using multiple reasoning paths per query alleviates this problem, but treating each path identically is suboptimal as paths vary widely in quality and suitability across tasks and models.
We propose Quality-filtered Routing with Cooperative Distillation
(QR-Distill), combining path quality filtering, conditional routing, and cooperative peer teaching. First, quality filtering retains only correct reasoning paths scored by an LLM-based evaluation. Second, conditional routing dynamically assigns paths tailored to each student's current learning state. Finally, cooperative peer teaching enables students to mutually distill diverse insights, addressing knowledge gaps and biases toward specific reasoning styles.
Experiments demonstrate QR-Distill's superiority over traditional single- and multi-path distillation methods. Ablation studies further highlight the importance of each component—quality filtering, conditional routing, and peer teaching—in effective knowledge transfer. Our code is available at~\href{https://github.com/LzyFischer/Distill}{https://github.com/LzyFischer/Distill}.

\end{abstract}

\section{Introduction}
\begin{figure}[t]
    \centering
    \includegraphics[width=\linewidth]{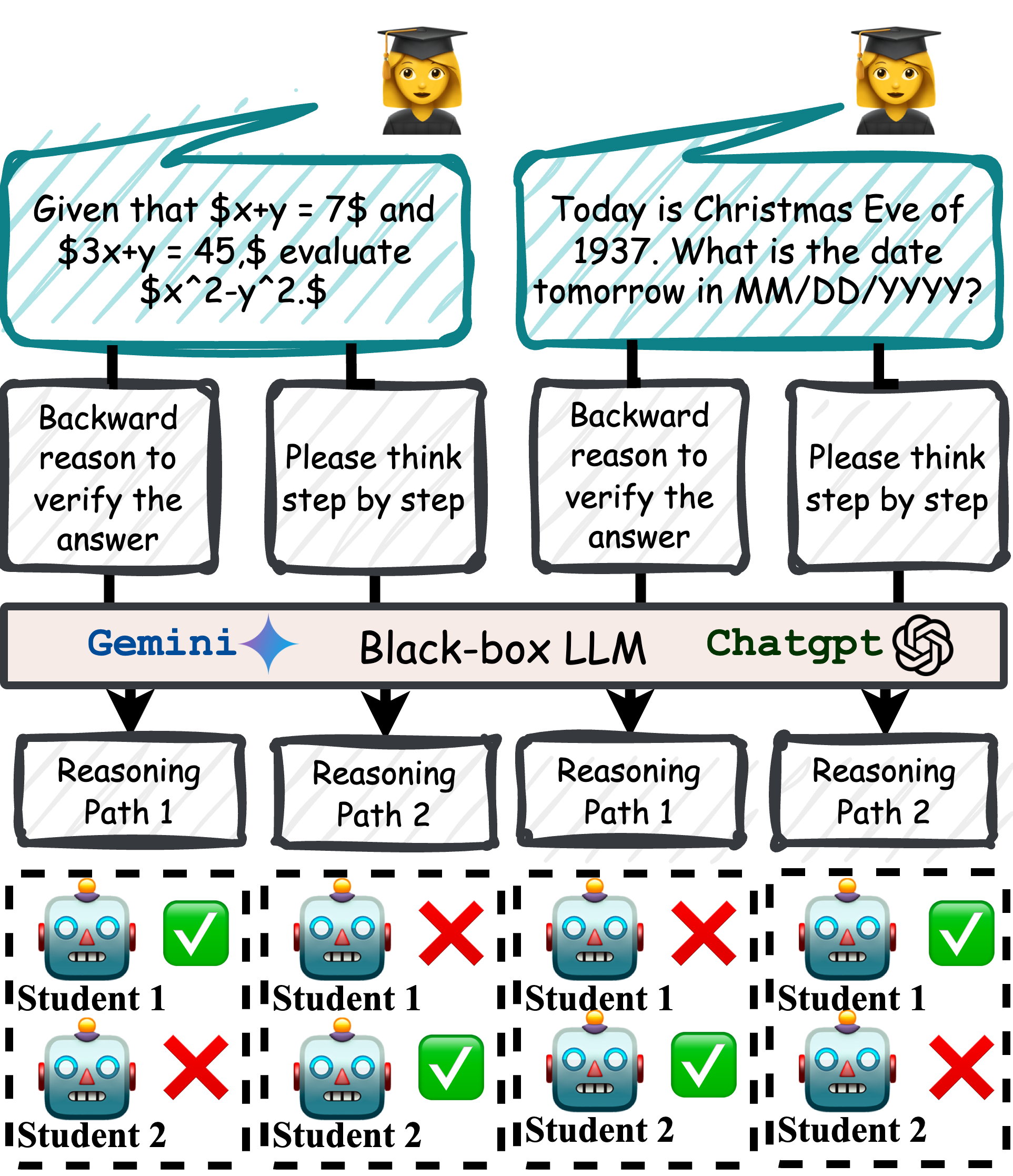}
    \caption{Distillation effectiveness of teacher-generated reasoning paths are path-, task-, and student-dependent. \includegraphics[height=1em]{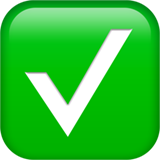} denotes effective, \includegraphics[height=1em]{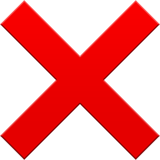} denotes ineffective distillation.}
    \label{fig:teaser}
    \vspace{-10pt}
\end{figure}
Recent scaling-law studies suggest that the reasoning abilities of large language models (LLMs) grows with model size and pre-training data~\cite{zhang2024scaling, yang2024harnessing, patil2024review, zhang2024scaling, lei2025harnessing}. Despite these advances, the high inference latency, memory demands, and licensing costs of proprietary black-box models limit their adoption in resource-constrained settings~\cite{agrawal2024taming, sun2024bbox, hong2023cyclealign}, thus ill-suited to many real-world deployments. Knowledge distillation provides a natural solution by training a compact and transparent student to replicate a powerful teacher~\cite{mcdonald2024reducing, xu2024survey, yang2024survey, muralidharan2024compact}, recovering most of the teacher’s competence while restoring efficiency and controllability.

Reproducing the teacher’s full reasoning ability remains challenging because conventional black-box distillation supervises students only at the token level~\cite{west2021symbolic, acharya2024survey, west2023novacomet}, which exposes only a narrow slice of the conditional distribution that underlies the teacher’s outputs. Empirical work shows that supervising on multiple chains of thought (CoTs) sampled for the same query can improve downstream accuracy~\cite{li2023symbolic, luo2025deconstructing}, suggesting that different reasoning trajectories capture complementary facets of the teacher’s problem-solving abilities and that aggregating them yields stronger learning signals than any single path alone. 

However, simply feeding every student all available paths is sub-optimal since the pedagogical value of reasoning paths is not universal. First, some traces arrive at incorrect conclusions~\cite{lyu2023faithful, trivedi2022interleaving} or embed spurious intermediate steps~\cite{he2021improving}, thus providing harmful teaching signals. Second, some reasoning paths are useful only for specific tasks or students, while irrelevant or even misleading for others, as shown in Figure~\ref{fig:teaser}. For example, program-style explanations often benefit algorithmic reasoning but add little value to routine arithmetic; long multi-hop chains help with complex commonsense puzzles but may overthink on questions that admit concise solutions~\cite{chen2024not}. Moreover, since student models differ in architecture, capacity, and pre-training data that leads to different learning abilities~\cite{turc2019well}, a reasoning path that aligns well with one learners can misguide another. As a result, Effective distillation requires path selection that is simultaneously quality-aware, task-aware, and student-aware.

We meet these requirements in two stages. (i) Quality filtering. We retain only paths whose final answers match ground truth labels, then score their internal reasoning with an LLM-as-judge, preserving the highest-rated traces. (ii) Conditional routing. For each query, a trainable router scores the surviving paths with respect to each student’s current state and selects the subset predicted to yield maximal learning gains.

Nevertheless, filtering narrows each student's view of the teacher’s knowledge again, risking a wider teacher–student gap and bias toward a limited set of reasoning styles. To close this gap, we introduce Quality-filtered Routing with Cooperative Distillation (QR-Distill), a cooperative framework in which multiple students train concurrently while acting as peer teachers. Each sample is processed in two passes: first in a teacher‑driven pass, where the router assigns the filtered paths to individual students, and then in a peer‑teaching pass, where a weighted ensemble of the students serves as a provisional teacher. A feature‑level mutual‑distillation loss channels information through this ensemble bottleneck, enabling learners to compensate for gaps in the others’ coverage, redistributing diverse insights obtained from the teacher's supervision.

We generate a broad, high-quality reasoning path pool by prompting an advanced black-box teacher with carefully designed variants, ensuring wide coverage of its solution space. Experiments on various benchmarks show that our framework consistently outperforms strong baselines that rely on either single-path distillation or multi-path distillation without routing. Ablation studies confirm that all components including quality filtering, conditional routing, and peer teaching contribute to the final gains, underscoring the value of path-aware selection and cooperative learning in distillation with multiple reasoning paths.

\section{Methodology}
Our method consists of four main components: (1) Reasoning Path Generation to augment training data, (2) Quality Filtering to eliminate incorrect paths, (3) Conditional Routing to assign reasoning paths to students adaptively, and (4) Mutual-Student Distillation to enable information exchange across student models, each elaborated below.

\subsection{Problem Setup}
Let $\mathcal{D} = \{(Q^{(i)}, A^{(i)})\}_{i=1}^{n}$ denote a reasoning dataset consisting of $n$ samples, where each sample consists of a question $Q^{(i)}$ and its corresponding ground-truth answer $A^{(i)}$. We assume black-box access to a teacher model $T$, meaning we can obtain outputs but not logits. Our goal is to train a smaller student model $s$ to improve its reasoning ability. During training, We augment $\mathcal{D}$ to obtain a new dataset $\mathcal{D}_{\text{aug}} = \{(Q^{(i)}, \mathcal{R}^{(i)})\}_{i=1}^{n}$, where each $\mathcal{R}^{(i)} = \{R_1^{(i)}, R_2^{(i)}, \dots, R_k^{(i)}\}$ is a set of $k$ diverse reasoning paths generated by a black-box teacher model $\mathcal{T}$. The student model $s$ is trained on $D_{\text{aug}}$. At test time, the student receives a simple instruction along with a question, similar to zero-shot prompting~\cite{kojima2022large}.

\subsection{Reasoning Path Generation}
To induce diversity in reasoning styles of multiple generated reasoning paths, we design and apply a set of prompting templates, each tailored to elicit a specific reasoning skill. The categories include:

\begin{itemize}
    \item \textbf{Vanilla Reasoning:} Standard prompts which encourage simple and linear reasoning.
    \item \textbf{Chain-of-Thought Reasoning:} Prompts to decompose the problem into multiple fine-grained reasoning steps~\cite{wei2022chain}.
    \item \textbf{Tree-of-Thought Reasoning:} Prompts to explore multiple solution paths before converging on a final answer~\cite{yao2023tree}.
    \item \textbf{Program-Based Reasoning:} Prompts to synthesize Python-like pseudocode to solve algorithmic problems~\cite{liu2024llms}.
    \item \textbf{Backward Reasoning:} Prompts to generate backward reasoning consistent with forward reasoning, simulating reverse-thinking of a problem~\cite{chen2024reverse}.
    \item \textbf{Fact-Retrieval Reasoning:} Prompts guiding the model to recall and retrieve relevant factual information before reasoning.
    
\end{itemize}

\begin{figure}
    \centering
    \includegraphics[width=\linewidth]{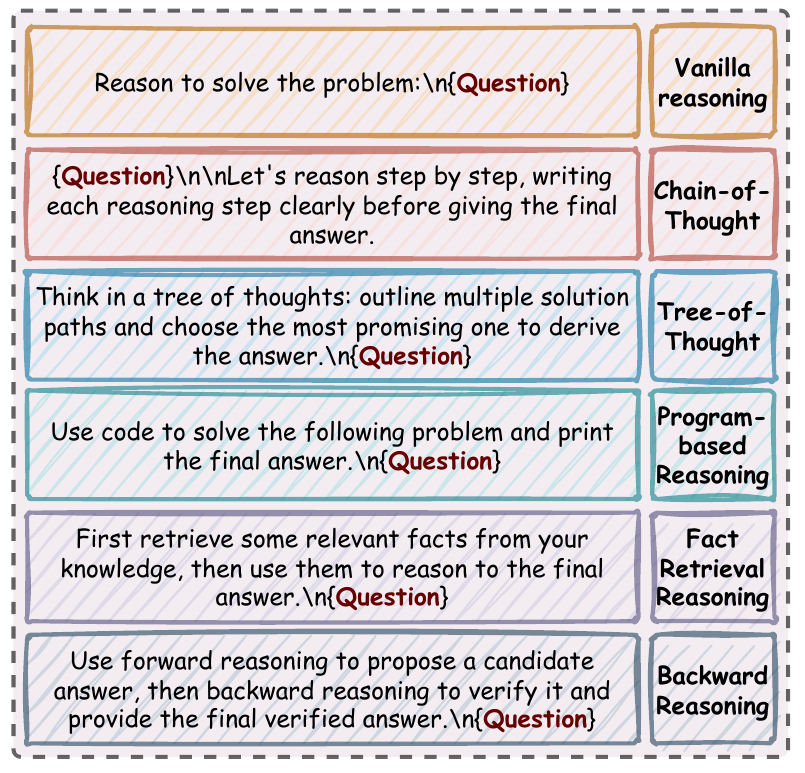}
    \caption{Prompt templates of different reasoning paths.}
    \label{fig:template}
\end{figure}

An example set of such prompt templates is illustrated in Figure~\ref{fig:template}.

\subsection{Quality Filtering}

Not all generated reasoning paths are equally informative or reliable for distillation. To ensure that the student model is trained on high-quality signals, we apply a two-stage filtering strategy that removes incorrect and misleading reasoning paths.

\paragraph{Step 1: Incorrect Answers Removal.} For each reasoning path $R_j^{(i)}$ generated for question $Q^{(i)}$, we extract the final predicted answer $\hat{A}_j^{(i)}$ and compare it against the ground-truth $A^{(i)}$. Paths for which $\hat{A}_j^{(i)} \neq A^{(i)}$ are discarded. This step ensures that only reasoning traces that lead to the correct solution are retained.

\paragraph{Step 2: Spurious Reasoning Removal.} The remaining paths are evaluated by a separate LLM-as-a-judge module $\mathcal{J}$, which is prompted to assess whether a path contains hallucinated or spurious intermediate steps. Only those marked as logically valid are retained. This yields a cleaned set $\widetilde{\mathcal{R}}^{(i)}$ of paths for each question.

\subsection{Conditional Routing}

While quality filtering removes clearly incorrect or spurious reasoning paths, it does so in a coarse and static manner. In practice, the usefulness of a reasoning path can vary depending on the query context and the specific student model. 
% A path that is helpful for one query or student may be irrelevant and even misleading for another. 
%
To enable more adaptive supervision, we introduce a \textit{conditional routing} mechanism that automatically assigns each reasoning path to one or more students. For each reasoning path $R_j^{(i)}$, we first extract a fixed representation using an encoder, i.e.,
\begin{equation}
    \mathbf{h}_j^{(i)} = \text{Enc}(\widetilde R_j^{(i)}) \in \mathbb{R}^d.
\end{equation}

Next, this representation is mapped to student-specific routing logits by a trainable router parameterized by an MLP, which are then processed via a Gumbel-Softmax to produce discrete but differentiable assignments, i.e.,
\begin{equation}
\boldsymbol{\alpha}_j^{(i)} = \text{GumbelSoftmax}(\text{MLP}(\mathbf{h}_j^{(i)})) \in \{0,1\}^S,
\end{equation}
where $\boldsymbol{\alpha}_j^{(i)}[s] = 1$ if reasoning path $\widetilde R_j^{(i)}$ is assigned to student $s$, and $0$ otherwise. $S$ denotes number of students involved during distillation. This allows the model to assign different reasoning paths to different students based on their compatibility, enabling adaptive supervision.

To prevent trivial cases such as always selecting all students or none, we apply an entropy-based regularization to promote balanced usage across students. Specifically, we average the routing assignment across all students and all reasoning paths and maximize its entropy, i.e.,
\begin{equation}
    \bar{\boldsymbol{\alpha}}^{(i)} = \frac{1}{S\cdot k} \sum_{j=1}^{k} \sum_{s=1}^{S} \boldsymbol{\alpha}_j^{(i)}[s],
\end{equation}
\begin{equation}
    \mathcal{L}_{\text{entropy}} = -\bar{\boldsymbol{\alpha}}^{(i)} \log \bar{\boldsymbol{\alpha}}^{(i)} - (1 - \bar{\boldsymbol{\alpha}}^{(i)}) \log (1 - \bar{\boldsymbol{\alpha}}^{(i)}).
\end{equation}

This regularization penalizes extreme routing decisions, thereby promoting informative and balanced supervision across students.

\begin{figure*}
    \centering
    \includegraphics[width=\linewidth]{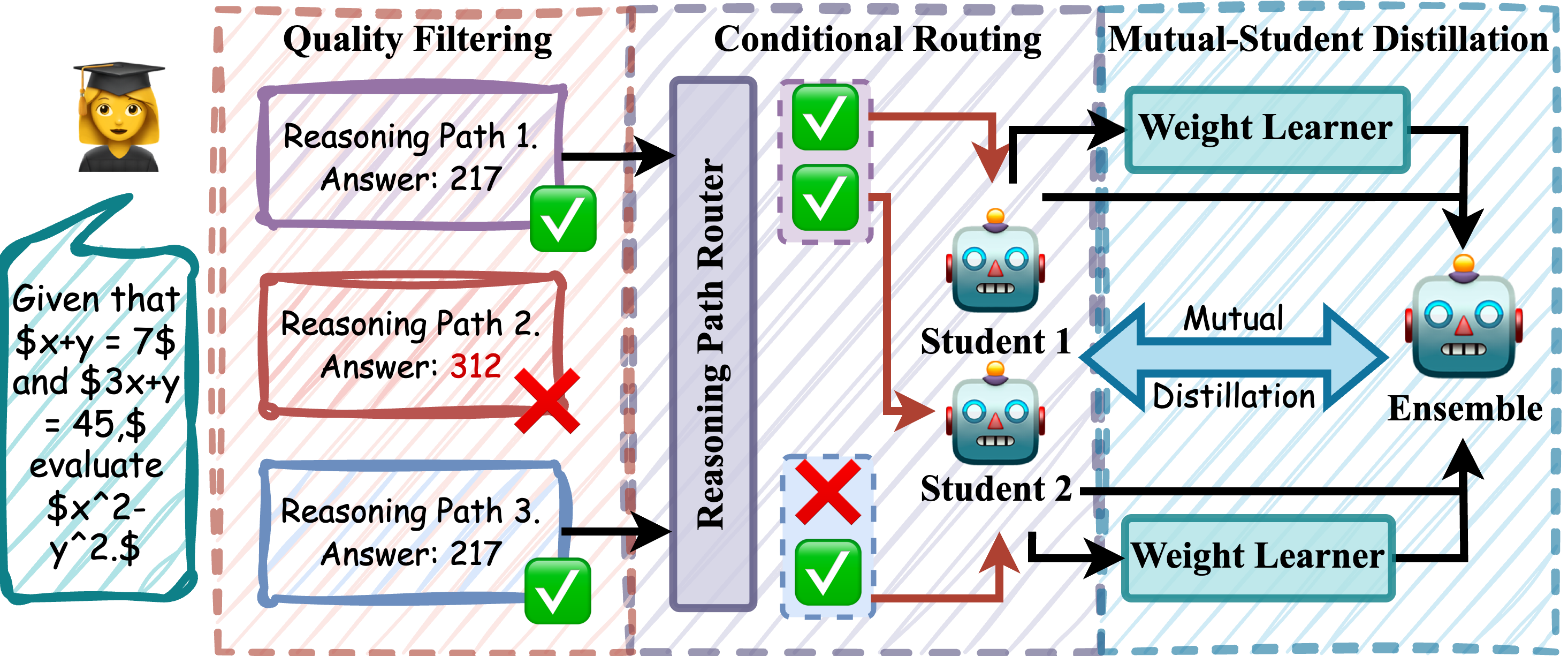}
    \caption{Overview of our framework, including \textbf{(1) Quality Filtering} that drops flawed chains-of-thought; \textbf{(2) Conditional Routing} that sends each reasoning path to the most suitable students for fine-tuning; \textbf{(3) Mutual-Student Distillation} that shares and refines learned insights of different students.}
    \label{fig:enter-label}
\end{figure*}
\subsection{Mutual-Student Distillation}

After filtering and routing, each student $S_s$ receives a subset of reasoning paths. However, isolated learning from limited reasoning styles may lead to narrow reasoning coverage and a persistent gap between students and the teacher. To mitigate this, we propose a \textit{mutual-student distillation} framework that allows students to learn from each other through internal representations of co-routed paths.

Let $\mathbf{z}_s^{(i,j)} \in \mathbb{R}^{T \times d}$ denote the last hidden states of student $s$ for path $\widetilde R_j^{(i)}$, where $T$ is the number of tokens. Each student projects their hidden states to a lower-dimensional shared space via a student-specific projection function, i.e.,

\begin{equation}
    \mathbf{\tilde{z}}_s^{(i,j)} = \text{Proj}_s(\mathbf{z}_s^{(i,j)}).
\end{equation}

We then compute a competence score $\gamma_s^{(i,j)}$ by averaging the projected hidden states across tokens and passing them through a linear regressor followed by a softmax over students, i.e.,
\begin{equation}
    \gamma_s^{(i,j)} = \text{softmax}_s\left( \mathbf{w}_s^\top \cdot \text{mean}_t(\mathbf{\tilde{z}}_s^{(i,j)}) \right),
\end{equation}

The scores are used to form a soft ensemble representation of the reasoning path, which includes knowledge from both students, i.e.,

\begin{equation}
    \mathbf{z}_{\text{ens}}^{(i,j)} = \sum_{s=1}^{S} \gamma_s^{(i,j)} \cdot \mathbf{\tilde{z}}_s^{(i,j)}.
\end{equation}

Each student then aligns its representation with the ensemble via a mean-squared error loss, i.e.,

\begin{equation}
    \mathcal{L}_{\text{mutual}} = \sum_{s=1}^{S} \sum_{i,j} \left\| \mathbf{\tilde{z}}_s^{(i,j)} - \mathbf{z}_{\text{ens}}^{(i,j)} \right\|_2^2.
\end{equation}

This mutual distillation allows each student to benefit from complementary knowledge learned by its peers, thereby reducing the gap between student and teacher.

\subsection{Training Objective}

The full objective function combines vanilla distillation losses, entropy regularization for the router, and mutual distillation losses:

\begin{equation}
    \mathcal{L} = \sum_{s=1}^{S} \mathcal{L}_{\text{distill}}^{(s)} + \lambda_1 \mathcal{L}_{\text{entropy}} + \lambda_2 \mathcal{L}_{\text{mutual}},
\end{equation}

where $\mathcal{L}_{\text{distill}}^{(s)}$ denotes supervised fine-tuning (SFT) loss for student $s$ on the reasoning paths assigned by the router. $\lambda_1$ and $\lambda_2$ control the relative importance of the other two losses.

\section{Experimental Setup}

\begin{table*}[t]
\centering
\resizebox{0.85\textwidth}{!}{%
\begin{tabular}{lcccccc}
\toprule
 \textbf{Methods} & \textbf{SQA} & \textbf{ARC} & \textbf{MATH} & \textbf{ANLI} & \textbf{Date} & \textbf{Avg} \\
\midrule
\multicolumn{7}{c}{\textbf{\textit{Gemini-1.5-Pro-001 (Teacher Model)}}}\\
\midrule
Zero-shot~\cite{kojima2022large}       & 77.39 & 91.51 & 55.90  & 70.12 & 80.00 & 79.76  \\
\midrule

\multicolumn{7}{c}{\textit{\textbf{Mistral-7B-Instruct}}} \\
\midrule
 Zero-shot~\cite{kojima2022large}   & 53.89 & 73.68 & 10.42 & 43.92 & 39.64&  44.31 \\
  \cmidrule(lr){2-7}
 SKD~\cite{li2023symbolic}     &   63.76  &74.66& 12.48  & 44.90 &48.50 & 48.86   \\
 Distill Step-by-Step~\cite{hsieh2023distilling}   &   64.19 &  75.32 & 11.54   & 44.42&  49.63 & 49.02\\
 Rephrase Question~\cite{yu2024natural}   & 65.07 & 74.51&  12.98& 43.58 & 45.51 & 48.33 \\
 Question Aug~\cite{li2024survey}    & 65.07 & 73.32 &13.64 &42.20 &47.21 & 48.29 \\
 \cmidrule(lr){2-7}
 Answer Aug~\cite{yu2024natural}  & 66.38 &  76.77&  14.78 & 45.01 & 49.12 & 50.41 \\
 RevTHINK~\cite{chen2024reverse} &  \textbf{70.97}  & 78.50 & 15.28  & 48.58 & 70.40 & 56.75 \\
  \cmidrule(lr){2-7}
 \textbf{QR-Distill (Ours)}     & 69.87  & \textbf{80.25} & \textbf{16.92}   &  \textbf{55.75} & \textbf{73.37} &  \textbf{59.23} \\
\midrule

\multicolumn{7}{c}{\textit{\textbf{Gemma-7B-Instruct}}} \\
\midrule
 Zero-shot~\cite{kojima2022large}    & 56.33 & 68.34 & 8.58   & 37.92 & 40.24 & 42.28 \\
  \cmidrule(lr){2-7}
 SKD~\cite{li2023symbolic}    &  56.77  & 73.29 & 16.86  & 45.42 & 59.62 & 50.39 \\
 Distill Step-by-Step~\cite{hsieh2023distilling} &   56.77  & 72.92 & 16.04  & 44.23 & 60.91 & 50.17\\
 Rephrase Question~\cite{yu2024natural} & 54.15 & 72.37 & 16.96 & 43.07 & 57.99 & 48.91 \\
 Question Aug~\cite{li2024survey}  & 55.10&  72.74 & 17.76   & 41.22 & 59.83 & 49.33 \\
\cmidrule(lr){2-7}
 Answer Aug~\cite{yu2024natural}  & 57.21  & 73.92 & 18.92  & 42.72 & 64.14 & 51.38 \\
 RevTHINK~\cite{chen2024reverse} & 64.19  & 75.09 & 19.96   & 47.36 & 66.27 & 54.57 \\
  \cmidrule(lr){2-7}
 \textbf{QR-Distill (Ours)}    &  \textbf{67.29} & \textbf{78.05} & \textbf{23.32}  & \textbf{51.50} & \textbf{79.29} &  \textbf{59.89} \\
\bottomrule
\end{tabular}
}
\caption{Performance comparison across five reasoning benchmarks with two students: \textit{Mistral-7B-Instruct} and \textit{Gemma-7B-Instruct}. Results are reported from prior work unless noted. Best values are bolded.}
\label{tab:main}
\end{table*}

\subsection{Backbone Models}

We use \texttt{Gemini-1.5-Pro-001}~\cite{team2024gemini} as the black-box teacher model $\mathcal{T}$, chosen for its strong reasoning performance across diverse domains. We train $S=2$ student models and instantiate them as \texttt{Mistral-7B-Instruct-v0.3} \cite{jiang2024mixtral} and \texttt{Gemma-7B-Instruct} \cite{team2024gemma}, both of which are widely-used open-weight instruction-tuned LLMs for distillation~\cite{chen2024reverse}. For encoding reasoning paths during routing, we use a pretrained \texttt{RoBERTa-base} model~\cite{liu2019roberta}.

\subsection{Training Details}

All students are fine-tuned using QLoRA~\cite{dettmers2023qlora} with rank 32. The learning rate is set to $5 \times 10^{-6}$ for Mistral and $2 \times 10^{-4}$ for Gemma, and remains consistent across all experiments.
Each student model is fine-tuned using the AdamW optimizer with a batch size of 8 per device. We train for 3 epochs on mathematical reasoning datasets (MATH, GSM8K) and 10 epochs on all other tasks. 

\subsection{Datasets}
We evaluate our method across diverse reasoning benchmarks spanning multiple domains, including \textbf{(1) Commonsense Reasoning:} StrategyQA~(SQA, \citet{geva2021did}) and ARC-Challenge~(ARC, \citet{clark2018think}); \textbf{(2) Mathematical Reasoning:} Math~\cite{hendrycks2021measuring}; \textbf{(3) Natural Language Inference:} ANLI~\cite{nie2019adversarial}; \textbf{(4) Logical Reasoning:} Date~\cite{srivastava2022beyond}. 

\vspace{-2pt}
\subsection{Baselines}
\vspace{-2pt}
We compare against three categories of baselines. \textbf{(1) Zero-shot:} Standard CoT prompting without fine-tuning~\cite{kojima2022large}. \textbf{Single-Path Distillation:} This includes \textit{(2) Symbolic Knowledge Distillation (SKD)}~\cite{li2023symbolic}, which trains on teacher-generated CoTs using next-token prediction, and \textit{(3) Distilling Step-by-Step}~\cite{hsieh2023distilling}, which adds supervision on both rationale and answer. We also include question-level augmentation methods: \textit{(4) Question Rephrasing}~\cite{yu2023metamath} and \textit{(5) Question Generation}~\cite{li2021stylized}. \textbf{Multi-Path Distillation:} These methods leverage multiple teacher-generated reasoning paths, including \textit{(6) Answer Augmentation}~\cite{yu2023metamath} and \textit{(7) Backward Reasoning Augmentation}~\cite{chen2024reverse}.

\section{Results and Analysis}
In this section, we aim to address four research questions. \textbf{RQ1}: How does \textsc{QR-Distill} compare with existing baselines? \textbf{RQ2}: What is the impact of each module inside \textsc{QR-Distill}? \textbf{RQ3}: How does the conditional router assign reasoning paths? \textbf{RQ4}: How does QR-Distill perform under varying training sample size?

\subsection{Main Results}

To address RQ1, we present our main results in Table~\ref{tab:main}. Overall, QR-Distill outperforms all baselines across datasets and models. Compared to the zero-shot performance of the student model, QR-Distill achieves an average improvement of $41.44\%$ with Mistral and $63.33\%$ with Gemma, indicating that knowledge learned from the teacher model can significantly enhance student performance on downstream reasoning tasks. When compared to baselines in which teachers provide only a single reasoning path for distillation, QR-Distill yields a substantial performance gain of $24.32\%$ on average, demonstrating that leveraging multiple reasoning paths leads to more effective student training. Against baselines that also use multiple reasoning paths but without our routing or collaborative mechanisms, QR-Distill still achieves up to $13.36\%$ improvement, which highlights the benefit of our path-aware routing and multi-student collaboration design in distilling diverse reasoning signals.

We also observe several noteworthy patterns. QR-Distill shows a larger performance boost for Gemma compared to Mistral across most datasets. Interestingly, on the Date dataset, Gemma even outperforms Mistral under QR-Distill, whereas it consistently underperforms in other baselines. This suggests that weaker student models benefit more from our method, likely due to the mutual distillation effect where Gemma learns useful patterns from its peer Mistral, which helps bridge the gap between Gemma and the black-box teacher.

Finally, we find that QR-Distill’s improvements are most pronounced on datasets where multi-path distillation baselines greatly outperform single-path ones, suggesting that QR-Distill can further unlock the potential of multiple reasoning paths.

\begin{table}[t]
\centering
\resizebox{0.9\columnwidth}{!}{%
\begin{tabular}{lcccc}
\toprule
 \textbf{Methods} & \textbf{ARC}  & \textbf{ANLI} & \textbf{Date} & \textbf{Avg} \\
\midrule
% \multicolumn{5}{c}{\textbf{\textit{Gemini-1.5-Pro-001 (Teacher Model)}}}\\
% \midrule
% Zero-shot      & 91.51 & 70.12 & 80.00  \\
% \midrule

\multicolumn{5}{c}{\textit{\textbf{Mistral-7B-Instruct}}} \\
\midrule
 \textit{w/o} QF   & 77.98   & 53.04 &  66.86 &  65.69 \\
 \textit{w/o} Route     &  78.07   & 59.00 & 72.78 &  69.95 \\
 \textit{w/o} Collab   &   75.38  &  \textbf{59.16}  & 72.19 & 68.91 \\
  \cmidrule(lr){2-5}
 \textbf{QR-Distill}     &  \textbf{80.25}  &  55.75 & \textbf{73.37} &  \textbf{69.79} \\
\midrule

\multicolumn{5}{c}{\textit{\textbf{Gemma-7B-Instruct}}} \\
\midrule
  \textit{w/o} QF   & 68.00   &  31.10  & 69.23 &  56.11 \\
 \textit{w/o} Route     &   75.19  & 30.17 & 78.10 & 61.15 \\
 \textit{w/o} Collab   &   77.88  &  46.33  & 76.33  & 66.85 \\
  \cmidrule(lr){2-5}
 \textbf{QR-Distill}     &  \textbf{78.05}   &  \textbf{51.50} & \textbf{79.29} &  \textbf{69.61} \\
\bottomrule
\end{tabular}
}
\caption{Ablation results on ARC, ANLI, and Date. Best values are bolded.}
\label{tab:ablation}
\end{table}

\begin{figure}[ht!]
    \centering
    \includegraphics[width=\linewidth]{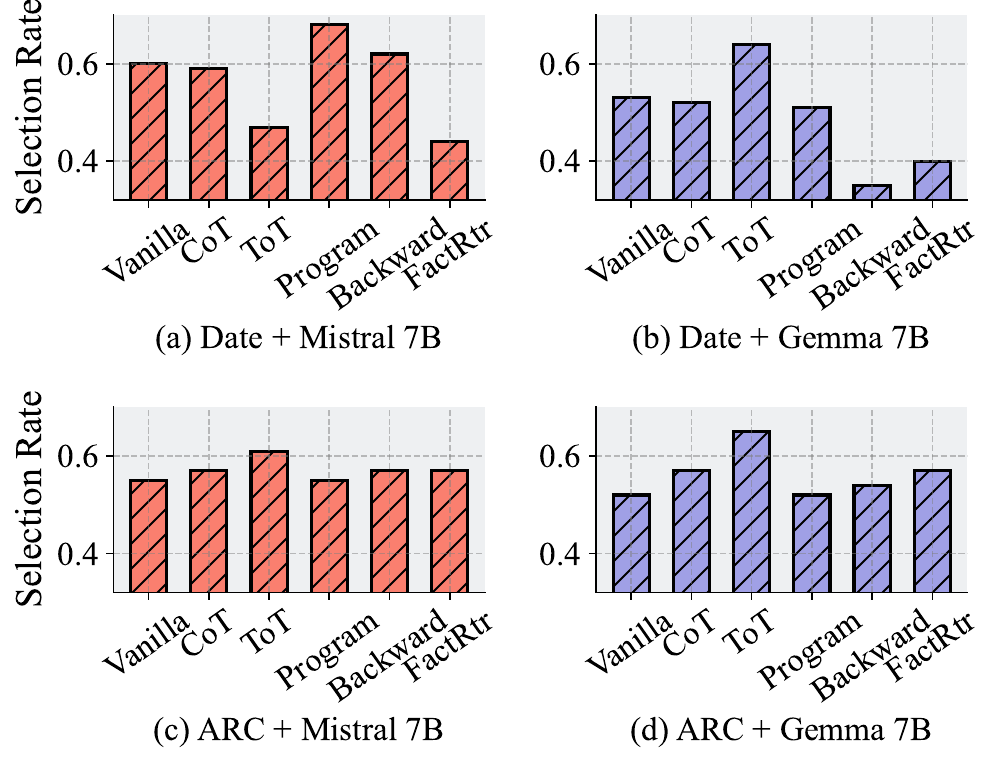}
    \caption{Routing selection rates across different dataset and  student model architectures.}
    \label{fig:router_dataset}
\end{figure}

\subsection{Ablation Study}
% To answer RQ2, we ablate from different modules and want to see how each of them contribute to the final performance. In the figure, we denote quality filtering as QF, Route as conditional routing, Collab as having mutual distillation between the two students. We have the following observations: (1) for most of the datasets, removing each module of QR-Distill will lead to degraded performance, which demonstrates that all of the modules can improve the distillation of small language models from the teacher. (2) quality filtering contribute the most to the final performance, which demonstrate that the reasoning paths with fault answers or intermediate steps can potentially do harm to the student model, worsen its hallucination. ANLI is influenced the most, which means the sensitivity of natural language inference tasks to the wrong reasoning paths. (3) Removing collab can do more harm to Gemma models, which verifies that weaker models have benefited more from the mutual distillation. and further demonstrated the effectiveness of our mutual distillation module. (4) Removing route can have better performance, which demonstrates that the importance of selecting compatible reasoning paths for ditillation.

To address RQ2, we conduct an ablation study by systematically removing different components of QR-Distill to assess their individual contributions. In the Table~\ref{tab:ablation}, we denote QF as Quality Filtering, Route as Conditional Routing, and Collab as Mutual-Student Distillation. Our observations are summarized as follows:
(1)	Across most datasets, removing any individual module results in performance degradation, suggesting that each component contributes to the overall distillation process. 
(2)	Among the three components, Quality Filtering appears to contribute the most consistently. This supports the hypothesis that filtering out low-quality reasoning paths particularly those with incorrect final answers or spurious intermediate steps can help reduce harmful supervision signals and mitigate potential hallucinations in the student models. This effect is especially pronounced on ANLI, suggesting that natural language inference tasks may be more sensitive to the quality of reasoning chains.
(2)	The Mutual Distillation module seems particularly beneficial for the Gemma student, as its removal results in more noticeable performance drops compared to Mistral. This aligns with our earlier observation that weaker models tend to benefit more from peer collaboration.
% (3)	Interestingly, removing the Routing mechanism sometimes leads to marginally better performance on certain datasets. This indicates that while routing reasoning paths to students based on compatibility is generally helpful, its current implementation may not always select the most effective path-student pairs. These results suggest potential room for improving the routing strategy to better align with student capabilities and task characteristics.

\begin{figure}[t]
    \centering
    \includegraphics[width=\linewidth]{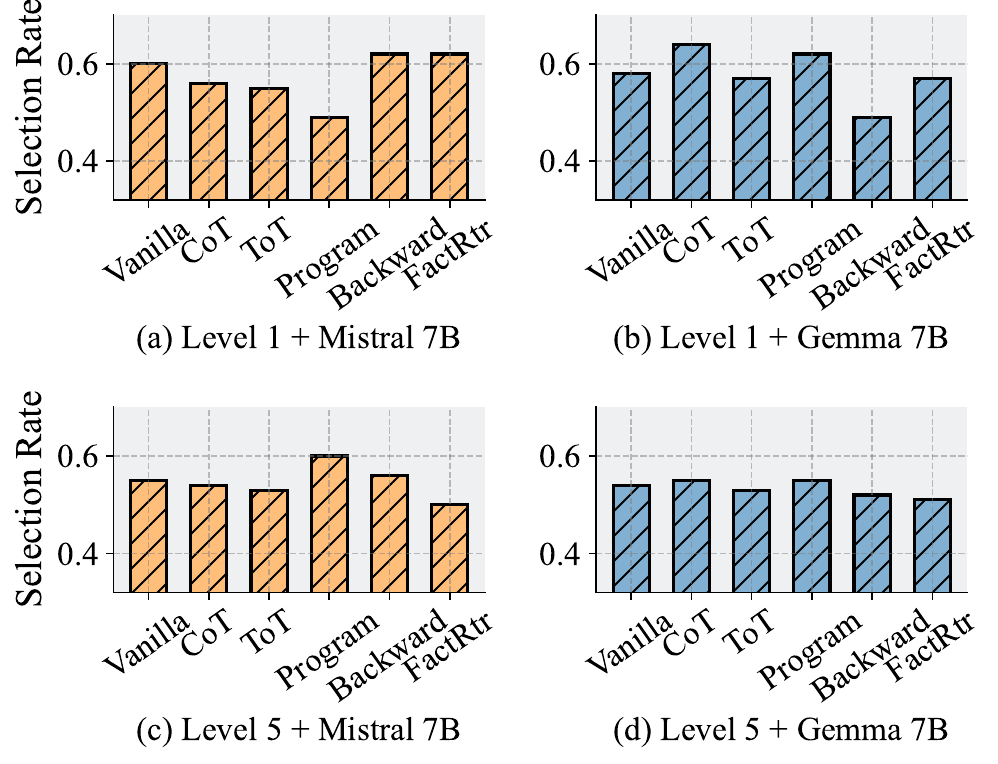}
    \caption{Routing selection rates across different question difficulty levels and student model architectures.}
    \label{fig:router_math}
\end{figure}

\subsection{Routing Analysis}

To answer RQ3, we analyze the routing decisions made for different reasoning paths across the two student models. Specifically, we investigate whether the \textbf{domain} and \textbf{difficulty} of questions influence routing behavior.
For the domain aspect, we compare routing choices across datasets. In Figure~\ref{fig:router_dataset}, \textit{CoT} denotes chain-of-thought, \textit{ToT} denotes tree-of-thought, \textit{program} refers to program-based reasoning, \textit{backward} denotes backward reasoning, and \textit{FactRtr} indicates fact-retrieval reasoning. We make the following observations:
(1) For the same dataset, the two students often select different reasoning paths, suggesting that compatibility between reasoning styles and model architecture can vary.
(2) For the same student, different datasets lead to different path preferences, indicating that question domain affects routing decisions.
(3) Fact-retrieval reasoning is favored on the ARC-Challenge dataset instead of the Date dataset, which aligns with our intuition that commonsense tasks rely more on factual recall than structured reasoning.
(4) A trade-off is observed between program-based and tree-of-thought reasoning, where when one is preferred, the other is often suppressed, suggesting a possible antagonistic relationship between these reasoning styles.

For question difficulty, we examine routing on the Math dataset at varying levels of complexity in Figure~\ref{fig:router_math}. We have the following observations:
(1) At the same difficulty level, different students favor different reasoning paths, further verifying the existence of student-reasoning path compatibility.
(2) Easier questions have higher selection rates, possibly reflecting a greater gap between student and teacher on more challenging questions.
(3) As question difficulty increases, differences in routing across reasoning paths diminish, suggesting a limitation in the students’ ability to effectively assess and select among reasoning strategies when facing complex problems.

% \subsection{Generalization Ability}

\subsection{Sample Efficiency}
% After showing that QR-Distill can outperform almost all baselines with the full training set, we want to answer RQ 4 by comparing whether QR-Distill can outperform SKD with varying portions of the training data. We select the portion $p \in \{0.3, 0.6, 1.0\}$. Across multiple reasoning tasks, QR-Distill consistently outperforms SKD at all levels of p, even surpassing SKD at p = 1.0 with only 10\% of the data on. Furthermore, while SKD stagnates with varying p, QR-Distill shows a clear upward trend in performance when p increases.

\begin{figure}[t!]
    \centering
    \includegraphics[width=\linewidth]{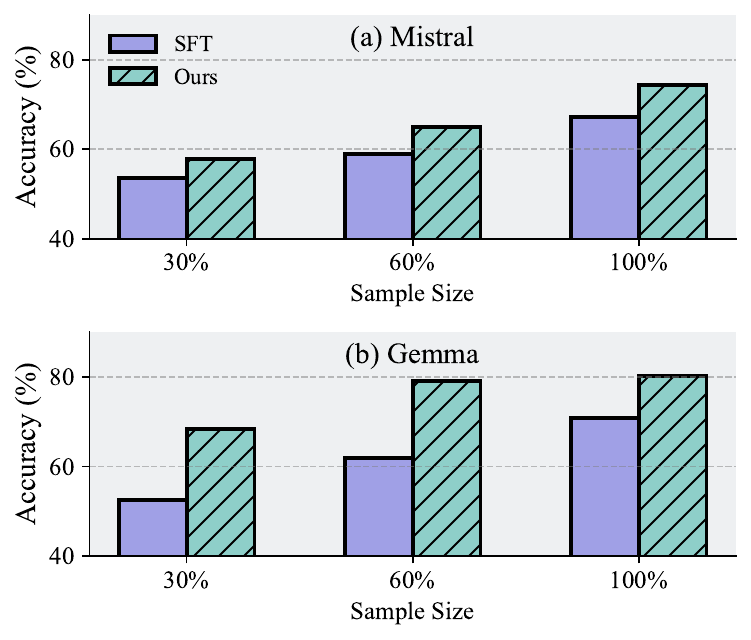}
    \caption{Comparison of QR-Distill and the SFT baseline with different sample sizes.}
    \label{fig:sample_efficiency}
\end{figure}

Having demonstrated the QR-Distill's performance on the full training set, we now address RQ4 by evaluating whether QR-Distill maintains its advantage under limited supervision. Specifically, we compare QR-Distill with SFT across varying ratios of the training data of Date dataset, as shown in Figure~\ref{fig:sample_efficiency}.
We can observe that QR-Distill consistently outperforms SFT at all training levels. Notably, QR-Distill is even comparable with SFT trained with $100\%$ data when using as little as $30\%$ data for Gemma, indicating better sample efficiency.

\section{Related Works}
\subsection{LLM Reasoning}
Recent advancements in LLMs have demonstrated significant capabilities in complex reasoning tasks~\cite{tan2025tuning, plaat2024reasoning, wang2024exploring, huang2022towards, yu2024natural, sun2023survey, ahn2024large, chen2025survey, tan2024glue, zhu2025collaborative, zheng2025corag}. A key factor behind this success is the use of advanced prompting techniques such as Chain-of-Thought (CoT) prompting~\cite{zhao2025chain, chu2023navigate, wei2022chain, lyu2023faithful, wei2025large} and Tree-of-Thought prompting~\cite{yao2023tree, long2023large, bi2024forest}. These methods encourage models to articulate reasoning explicitly, enhancing their ability to solve intricate problems. Building on CoT approaches, researchers have explored various strategies to further exploit the diversity and richness of multiple reasoning paths~\cite{naik2023diversity, chen2023universal, wang2024understanding}. For instance, Self-Consistency employs multiple reasoning samples from the same prompt, aggregating them via majority voting to improve answer reliability~\cite{wang2022self, chen2023two, liang2024internal, ahmed2023better, tan2025prospect, chen2025maple, yuan2025tracing, li2024political, chen2025from}.

Despite these improvements, existing strategies utilizing multiple reasoning paths largely focus on aggregating reasoning paths post-generation without adequately addressing the selective utilization of reasoning paths~\cite{yin2024aggregation, wang2024understanding, fang2024karpa}. Most approaches indiscriminately combine reasoning samples, which risks incorporating redundant or low-quality rationales~\cite{xu2023rethinking, wang2024qcrd,tong2024can}, potentially limiting model efficacy. A critical yet under-explored direction involves systematically identifying and selecting reasoning paths based on their quality, relevance, and compatibility with specific tasks and model characteristics.

\subsection{Knowledge Distillation}
Knowledge distillation (KD) aims to transfer knowledge from powerful but cumbersome teacher models to smaller student models~\cite{li2024contextualization,gou2021knowledge, hinton2015distilling, park2019relational, chen2021distilling}. Traditional KD approaches typically align the student's predictive distributions closely with those of the teacher, often requiring internal access to the teacher’s parameters~\cite{tan2024interpreting,zhao2022decoupled, cho2019efficacy, kim2016sequence, gu2023minillm}. However, such methods become impractical for proprietary and black-box LLMs~\cite{xu2024survey, yang2024survey, hong2023cyclealign}, motivating the exploration of distillation methods that rely on token-level model outputs.

Recently, symbolic distillation techniques have emerged, which leverage explicit rationales or symbolic outputs from large-scale teacher models without requiring internal access~\cite{acharya2024survey, west2021symbolic, li2023symbolic}. \citet{hsieh2023distilling} demonstrated that the utility of rationales in the distillation step by step can improve the performance and improve sample efficiency. In addition, \citet{jiang2023lion} propose a teacher-feedback mechanism where LLM-generated rationales for challenging examples guide student models.

Despite their effectiveness, these symbolic distillation approaches frequently employ a single reasoning path per query, thus inadequately capturing the teacher’s comprehensive reasoning capabilities. Consequently, recent efforts have explored multi-path distillation, integrating diverse CoT samples to enhance student performance~\cite{zhang2025questfor,chen2023mcc, chen2024reverse, li2023symbolic}. Nonetheless, most of these studies lack a rigorous selection mechanism for reasoning paths, risking the inclusion of suboptimal or irrelevant rationales, thus hindering the potential benefits. In addition, none of existing methods utilize the collaboration of students to improve the distillation of multiple reasoning paths.

\subsection{Multi-Agent Collaboration}
Multi-agent collaborative frameworks have demonstrated notable improvements in complex reasoning and problem-solving tasks by harnessing collective intelligence~\cite{tran2025multi, hong2023metagpt, talebirad2023multi, chen2023agentverse, li2023theory, li2024survey, zhang2025beyond, zhou2024teaching, lee2024mentor}. This is achieved by combining diverse perspectives and complementary capabilities to enhance overall performance. Through mechanisms such as information sharing~\cite{han2024llm}, joint decision-making~\cite{sun2024llm}, and iterative refinement~\cite{chen2024optima}, collaborative approaches consistently outperform isolated single-agent models.

Despite the advantages of collaborative frameworks, integrating these principles explicitly within knowledge distillation is relatively unexplored. Our approach uniquely combines collaboration of multiple student models with selective distillation, leveraging inter-agent cooperation to enhance reasoning path selection and learning, thereby addressing critical gaps identified in prior research.

\section{Conclusion}
We propose QR-Distill, a novel framework that addresses the varied suitability of multiple reasoning paths across tasks and student models. QR-Distill integrates three key components: (1) \textbf{Quality Filtering} to retain only high-quality, correct reasoning paths using an LLM-based evaluator; (2) \textbf{Conditional Routing} to adaptively assign paths to students based on their current learning state; and (3) \textbf{Mutual-Student Distillation} to enable mutual knowledge transfer among students, mitigating reasoning style bias and teacher-student gaps. Extensive experiments confirm the effectiveness of our approach in improving multi-path distillation.

\section*{Limitations}
\textbf{Limited number of student models.}
Due to constraints in computational resources, we conduct experiments using only two student models. While this setup already demonstrates the benefits of collaborative learning, increasing the number of collaborative students holds huge potential for further performance gains.

\noindent\textbf{Single teacher model.}
All reasoning paths in this work are generated using the Gemini-1.5 model. Although Gemini is a strong teacher, including outputs from additional teacher models such as GPT may expose students to a broader range of reasoning styles and improve generalization.

\noindent\textbf{Restricted diversity of reasoning prompts.}
We employ a predefined set of prompt templates to induce different reasoning styles. Exploring a wider set of reasoning path types could further enrich training signals and enhance the effectiveness of our distillation framework.

\section*{Ethics Statement}

Our work focuses on developing an effective distilling framework using publicly available datasets and pretrained LLMs. While acknowledging the need for responsible usage of the proposed method, we do not foresee major negative societal impacts.

\section*{Acknowledgments}
This work is supported in part by the National Science Foundation (NSF) under grants IIS-2006844, IIS-2144209, IIS-2223769, CNS-2154962, BCS-2228534, and CMMI-2411248; the Office of Naval Research (ONR) under grant N000142412636; and the Commonwealth Cyber Initiative (CCI) under grant VV-1Q24-011, and the gift funding from Netflix and Snap.

\bibliography{acl_latex}
\begin{table*}[t]
\centering
\small
\begin{tabular}{lcccccc}
\toprule
\textbf{Model} & \textbf{SQA} & \textbf{ARC} & \textbf{MATH} & \textbf{ANLI} & \textbf{Date} & \textbf{Avg} \\
\midrule
Mistral-7B-Instruct & 06:07 & 04:08 & 19:37 & 06:28 & 00:41 & 07:24 \\
Gemma-7B-Instruct & 07:50 & 04:59 & 24:45 & 08:00 & 00:54 & 09:18 \\
Sum & 13:57 & 09:07 & 44:22 & 14:28 & 01:35 & 16:42 \\
QR-Distill & \textbf{09:49} & \textbf{05:44} & \textbf{32:15} & \textbf{09:05} & \textbf{00:59} & \textbf{11:34} \\
\bottomrule
\end{tabular}
\caption{Training time per epoch (minutes). QR-Distill achieves efficiency gains via parallel supervision.}
\label{tab:efficiency}
\end{table*}
\newpage
\appendix

\section{Additional Experiments}
\paragraph{Training Efficiency.}  
To further examine computational efficiency, we compare the training time of QR-Distill with that of training individual student models separately. Since QR-Distill jointly supervises multiple students in a single run, it benefits from parallel supervision, whereas single-model baselines process only one rationale per sample. As shown in Table~\ref{tab:efficiency}, this design leads to reduced training time per epoch.

\paragraph{Smaller \& Different Sized Models.}  
To evaluate QR-Distill beyond 7B-scale students, we also conduct experiments with both smaller and differently sized models. This setting examines whether the proposed framework can still provide benefits when applied to lightweight architectures. Results on several datasets are summarized in Table~\ref{tab:small_models}.

\begin{table}[h]
\centering
\small
\begin{tabular}{lccc}
\toprule
\textbf{Model} & \textbf{ANLI} & \textbf{ARC} & \textbf{Date} \\
\midrule
TinyLLaMA & 2.33 & 23.50 & 15.38 \\
TinyLLaMA-1.1B-QR-Distill & \textbf{21.25} & \textbf{32.42} & \textbf{20.71} \\
\midrule
Qwen2.5-3B & 14.26 & 73.50 & 63.31 \\
Qwen2.5-3B-QR-Distill & \textbf{30.83} & \textbf{81.48} & \textbf{74.56} \\
\bottomrule
\end{tabular}
\caption{Preliminary results on smaller and differently sized student models.}
\label{tab:small_models}
\end{table}

\paragraph{Resource-Constrained Settings.}  
To explore QR-Distill under resource-constrained conditions, we conduct an experiment where one student is frozen while the other continues training. Table~\ref{tab:frozen} shows that the target student can still achieve lower performance in this setting, suggesting that static guidance from an untrainable peer may hinder effective knowledge transfer.

\begin{table}[h]
\centering
\small
\begin{tabular}{lcc}
\toprule
\textbf{Model} & \textbf{Frozen-Another} & \textbf{QR-Distill} \\
\midrule
Mistral & 66.27 & 73.37 \\
Gemma & 76.97 & 79.29 \\
\bottomrule
\end{tabular}
\caption{Results on the Date dataset when freezing one student. Freezing leads to degraded performance.}
\label{tab:frozen}
\end{table}

% \section{Appendix}
% \label{sec:appendix}

% This is an appendix.

\end{document}